\documentclass{article}

\usepackage{arxiv}

\usepackage[utf8]{inputenc}  
\usepackage[T1]{fontenc}     
\usepackage{lmodern}         
\usepackage{microtype}       
\usepackage[scaled]{helvet} 
\usepackage[english]{babel}  
\usepackage{graphicx}  

\usepackage{amsmath,amssymb,amsfonts}
\usepackage{algorithm, algorithmic}

\usepackage[bookmarks=false]{hyperref}            
\usepackage{comment}         

\graphicspath{{figures/}}

\title{Improving Data Quality with Training Dynamics of Gradient Boosting Decision Trees}
\renewcommand{\shorttitle}{Improving Data Quality with Training Dynamics of Gradient Boosting Decision Trees}

\author{Moacir A. Ponti\thanks{M. Ponti is also with ICMC/Universidade de S\~ao Paulo, S\~ao Carlos-SP, Brazil}, Lucas de Angelis Oliveira\\
	Mercado Livre\\
	Osasco, Brazil\\
\url{moacir.ponti@mercadolibre.com} \\
	\And
	Valentina Garcia\\
	Mercado Libre\\
	Medellín, Colombia\\
	\And
        Mathias Esteban\\
        Mercado Libre \\
Montevideo Uruguay\
	\And
	Juan Martín Román, Luis Argerich\\
	Mercado Libre\\
	Buenos Aires, Argentina
}
\date{}

\begin{document}
\maketitle

\begin{abstract}
Real world datasets contain incorrectly labeled instances that hamper the performance of the model and, in particular, the ability to generalize out of distribution. Also, each example might have different contribution towards learning. This motivates studies to better understanding of the role of data instances with respect to their contribution in good metrics in models. In this paper we propose a method based on metrics computed from training dynamics of Gradient Boosting Decision Trees (GBDTs) to assess the behavior of each training example. We focus on datasets containing mostly tabular or structured data, for which the use of Decision Trees ensembles are still the state-of-the-art in terms of performance. Our methods achieved the best results overall when compared with confident learning, direct heuristics and a robust boosting algorithm. We show results on detecting noisy labels in order clean datasets, improving models' metrics in synthetic and real public datasets, as well as on a industry case in which we deployed a model based on the proposed solution. 
\end{abstract}



\section{Introduction}
Investigating data quality is paramount to allow business analytics and data science teams to extract useful knowledge from databases. A business rule may be incorrectly defined, or unrealistic conclusions may be drawn from bad data. Machine Learning models may output useless scores, and Data Science techniques may provide wrong information for decision support in this context. Therefore, it is important to be able to assess the quality of training data~\cite{jain2020overview, smith2015potential}. 

Datasets for learning models can grow fast due to the possibility of leveraging data from the Internet, crowdsourcing of data in the case of academia, or storing transactions and information of business into data lakes in the case of industry. However, such sources are prone to noise, in particular when it comes to annotations~\cite{johnson2022survey}. Even benchmark datasets contain incorrectly labeled instances that affect the performance of the model and, in particular, the ability to generalize out of distribution~\cite{ekambaram2017finding, pulastya2021assessing}. In this context, while Machine Learning theory often shows benefits of having large quantities of data in order to improve generalization of supervised models, usually via the Law of Large Numbers~\cite{mello2018machine}, it does not directly addresses the case of data with high noise ratio.

In fact, different examples might not contribute equally towards learning~\cite{vodrahalli2018all, sorscher2022beyond}. This motivates studies to better understand the role of data instances with respect to their contribution in obtaining good metrics. Instance hardness may be a way towards this idea~\cite{zhou2020curriculum}. However, more than identifying how hard a given example is for the task at hand, we believe there is significant benefit in segmenting the dataset into examples that are useful to discover patterns, from those useless for knowledge discovery~\cite{hao2022model, saha2014data, frenay2013classification}. Trustworthy data are those with correct labels, ranging from typical examples that are easy to learn, ambiguous or borderline instances which may require a more complex model to allow learning, and atypical (or rare) that are hard-to-learn. 

Therefore, in this paper we propose a method based on metrics computed from training dynamics of Gradient Boosting Decision Trees (GBDTs) to assess the behavior of each training example. In particular, it uses either XGBoost~\cite{chen2016xgboost} or LightGBM~\cite{ke2017lightgbm} as base models. Our algorithm is based on the Dataset Cartography idea, originally proposed for Neural Networks in the context of natural language processing datasets~\cite{swayamdipta2020dataset}. In contrast, we focus on datasets containing mostly tabular or structured data, for which Decision Trees ensembles are the state-of-the-art in terms of performance, classification metrics, as well as interpretability~\cite{shwartz2022tabular}. Also~\cite{swayamdipta2020dataset} devote their main efforts to investigate the use of ambiguous examples to improve generalization and only briefly to mislabeled examples. In this study we instead focus on detecting noisy labels in order to either remove it or relabel it to improve models' metrics.

Our contributions are as follows:
\begin{enumerate}
    \item We are the first to introduce training dynamics metrics for dataset instances, a.k.a. Dataset Cartography, using ensembles of boosted decision trees (or GBDTs);
    \item Use the method as part of the pipeline to deploy a model in production used to classify forbidden items in a Marketplace platform and show guidelines for users that may benefit from the practices shown in our applied data science paper;
    \item Propose a novel algorithm that uses the computed training dynamics metrics, in particular a product between correctness and confidence, in conjunction with LightGBM iterative instance weights to improve noisy label detection;
    \item By investigating both Noise Completely At Random (NCAR) and noisy not at random (NNAR), we show that removing mislabeled instances may improve performance of models, outperforming previous work in many scenarios, including real, synthetic and a productive dataset.
\end{enumerate}

\section{Related Work}

Previous work includes approaches to score dataset instances using confidence~\cite{hovy2013learning} and metrics of hardness~\cite{lorena2019complex}. Beyond measuring confidence or hardness, the field known as ``confident learning''~\cite{northcutt2021confident} intents to address the issue of uncertainty in data labels during neural network training. Some important conclusions were drawn in this scenario for multiclass problems, in particular: (i) that label noise is class-conditional~\cite{angluin1988learning}, e.g. in an natural image scenario a \textit{dog} is more likely to be misclassified as \textit{wolf} than as \textit{airplane}; (ii) that joint distribution between given (noisy) labels and unknown (true) labels can be achieved via a series of approaches: pruning, counting and ranking. According to~\cite{northcutt2021confident}, prune is to search for label errors, for example via loss-reweighing to avoid iterative re-labeling~\cite{chen2019understanding, patrini2016loss}, or using unlabeled data to prune labeled datasets~\cite{sorscher2022beyond}. Count is to train on clean data in order to avoid propagating error in learned models~\cite{natarajan2013learning}. Then, to rank examples to use during training as in curriculum learning~\cite{zhou2020curriculum}.

Using learning or training dynamics for neural networks models was shown to be useful to identify quality of instances. For example, comparing score values with its highest non-assigned class~\cite{pleiss2020identifying} or instances with low loss values~\cite{shen2019learning}. Understand which instances represent simpler patterns and are easy-to-learn~\cite{liu2020early}, as well as those that are easily forgotten~\cite{toneva2018empirical} (misclassified) in a later epoch. Such studies show that deep networks are biased towards learning easier examples faster during training. In this context, making sure the deep network memorizes rare and ambiguous instances, while avoiding memorization of easy ones, lead to better generalization~\cite{feldman2020does,swayamdipta2020dataset, li2019repair}. Noise in both training and testing data led to practical limits in performance metrics requiring use of novel training approaches~\cite{ponti2021training}.  This is important in the context of neural networks since usually such models require an order of magnitude more data in order to improve metrics in $3-2\%$~\cite{sorscher2022beyond}.

AdaBoost versions designed to be robust to noise have been proposed such as LogitBoost~\cite{friedman2000additive} and later BrownBoost~\cite{freund2001adaptive}. Also in~\cite{ratsch2000robust} boosting is defined as a margin maximization problem, inspired by Statistical Learning Theory, and slack variables are introduced to allow for a soft version which allow a fraction of instances to lie inside the margin. More than a decade after such studies, Gradient Boosting Decision Trees (GBDTs) were proposed and dominated the class of tabular problems, excelling in both performance and speed~\cite{shwartz2022tabular}. The most recent one, LightGBM~\cite{ke2017lightgbm}, is currently the standard choice in this sense. Decision trees are also shown to be robust to low label noise~\cite{ghosh2017robustness}, making it a feasible model to investigate under significant noise regimes. 

While more recent work addresses issues closely related to deep neural networks and large scale image and text datasets, studies on datasets containing tabular or structured data are still to be conducted~\cite{renggli2023automatic}. The concepts defined in the next section were defined before in different studies such as~\cite{smith2015potential, swayamdipta2020dataset}, or focus on AdaBoost using an arbitrary or manual choice as a threshold for noise robustness~\cite{karmaker2006boosting, friedman2000additive}. In this paper we define training dynamic metrics for the first time for GBTDs, also we were the first to use a combination of training dynamic metrics, as well as propose an automatic algorithm to define a threshold to assess label noise 

\begin{figure}[htbp]
\centerline{\includegraphics[width=0.75\linewidth]{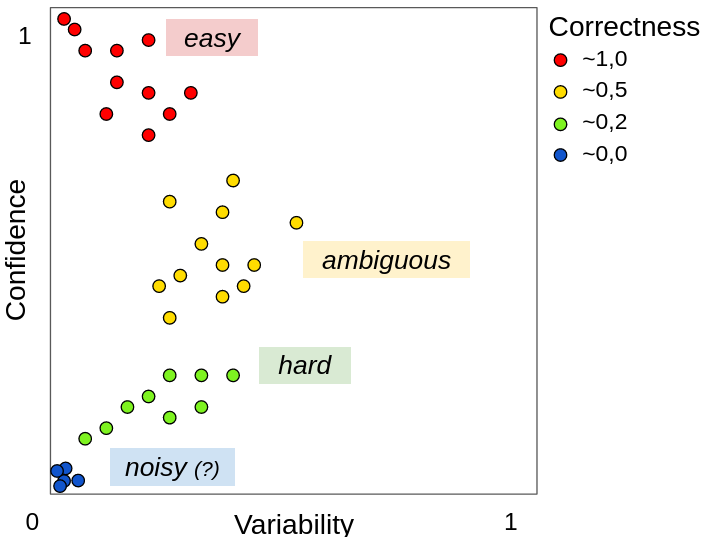}}
\caption{Dataset cartography illustration based on training dynamics: average confidence, variability and correctness, allowing to map the instances to how the model evolved to estimate the outputs along iterations and classify points into easy, ambiguous, hard and even noisy.}
\label{fig:cartography_scheme}
\end{figure}

\section{Dataset Cartography using Training Dynamics of Decision Trees}

In the context of Boosting-based Decision Trees ensembles, the training dynamics are given by a sequence of trees, each learned by using as input weights for the misclassified instances in the previous iteration. For each iteration $i$ of a GBDT model $i$ trees (estimators) are used to compute the probabilities/scores for each class and all instances of the dataset. An advantage of using such ensembles, such as LightGBM and XGBoost, is that we are able to compute scores at any iteration using an already trained model, without the need to retrain it from scratch, as in neural networks.

We define as $p^*_{(i)}(y^*_j|x_j)$ the score the model predicted for the classes of each instance $j$, where the input is $x_j$ and its training label $y^*_j$. The predicted label is $\hat{y}_j$. Note that $\hat{y}_j$ (predicted by the model) may be equal or different than $y^*_j$ (training label). This method is supervised, requiring $y^*_j$, and therefore allows to assess only training instances.

The following training dynamics statistics are computed for each instance:
\begin{itemize}
    \item \textbf{Confidence}: the average score for the true label $y^*_j$ across all iterations:
    \begin{equation*}
        \mu_j = \frac{1}{T} \sum_{i=1}^{T} p^*_{(i)}(y^*_j|x_j),
    \end{equation*}
    where $p^*_{(i)}$ is the model's score at iteration $i$ relative to the true label (not the highest score estimated by the model);
    \item \textbf{Correctness}: the percentage of iterations for which the model correctly labels $x_j$:
    \begin{equation*}
        c_j = \frac{1}{T} \sum_{i=1}^{T} (\hat{y}_j = y^*_j),
    \end{equation*}
    \item \textbf{Variability}: the standard deviation of $p^*_{(i)}$ across iterations:
    \begin{equation*}
        \sigma_j = \sqrt{ \frac{\sum_{i=1}^{T} (p^*_{(i)}(y^*_j|x_j) - \mu_j)^2}{T}}.
    \end{equation*}   
\end{itemize}

The metrics are in the range $[0,1]$. The name ``dataset cartography'' comes from a visualization of such metrics as proposed by~\cite{swayamdipta2020dataset}, and illustrated in Figure~\ref{fig:cartography_scheme}.  Algorithm~\ref{algo:cartography} details how to compute such metrics using GBDTs, given a trained model $h(.,.)$ for which it is possible to get the output for an iteration $i$, and the dataset used to train this model $X,Y$.

\begin{algorithm}
\caption{Dataset Cartography: Training Dynamic Metrics for GBDTs}
\label{algo:cartography}
\begin{algorithmic}
\REQUIRE Training set: $X, Y$
\REQUIRE Trained model: $h(., .)$
\REQUIRE Number of estimators/iterations: $T$
\STATE $\hat{S} \gets \emptyset$ \COMMENT{initialize set of scores}
\STATE $\hat{Y} \gets \emptyset$ \COMMENT{initialize set of labels}
\STATE \COMMENT{for each iteration}
\FOR {$i=0$ \TO $T-1$}
\STATE $\hat{s}_i \gets h(X,i)$ \COMMENT{predictions for all classes} 
\STATE $\hat{y}_i \gets \arg \max(\hat{s}_i)$ \COMMENT{predicted labels of all items}

\STATE $\hat{S} \gets \hat{S} \cup p^*_{(i)}(y^*|x)$ \COMMENT{scores for training labels at $i$}
\STATE $\hat{Y} \gets \hat{Y} \cup \hat{y}_i$ \COMMENT{store predicted labels at $i$}
\ENDFOR
\STATE $\mu \gets \operatorname{mean}(\hat{S})$ \COMMENT{average of scores for training labels}
\STATE $\sigma \gets \operatorname{std}(\hat{S})$ \COMMENT{spread of scores}
\STATE $c \gets \operatorname{sum}(\hat{Y}==Y)/T$ \COMMENT{percentage of correct predictions}
\RETURN $\mu, \sigma, c$
\end{algorithmic}
\end{algorithm}
Possible interpretations are as follows. A \textit{high confidence} example can be considered easy for the model to learn. An example for which the model assigns the same label has \textit{low variability}, which associated with high confidence makes the example even easier. In contrast, \textit{high variability} means different subset of trees provide different scores for an instance, indicating harder instances, that are usually associated to complex patterns, rare or atypical ones. \textit{High correctness} are associated to easy instances, but such instances may have confidence ranging from 1.0 to 0.5), while \textit{near zero correctness} indicates an instance that the model cannot learn.

\section{Improving Dataset Quality}

Each metric (confidence, variability, correctness) may be applied in different ways to assess the dataset. In this paper, we propose a series of approaches to use training dynamics statistics in order to improve quality of datasets. In particular we address the problem of detecting pathological examples. We define such examples those with wrong label and that potentially harm the model's performance. The proposed methods, from the simpler to the most complex, are as follows: (i) using a combination of confidence and correctness to remove useless instances by threshold, (ii) iteratively learn pathological examples in order to separate them from the good ones by combining the gradient boosting instance weights and the training statistics. 

\subsection{Threshold based on product between confidence and correctness}

While previous works often uses confidence of each example along epochs or iterations, or filter instances by correctness~\cite{smith2015potential}, we multiply the values of confidence and correctness, i.e. $m_j = c_j \cdot \mu_j$ for each instance $j$ and decide whether to keep or remove a given instance from the training set. 

\begin{figure}[htbp]
\centerline{\includegraphics[width=0.7\linewidth]{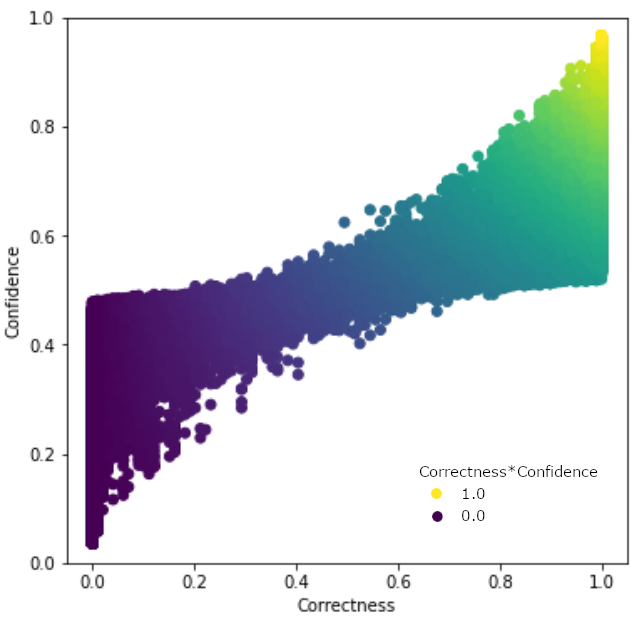}}
\caption{Correctness versus Confidence for each instance, as well as their multiplication (coded in different colors).}
\label{fig:correc_confid_p}
\end{figure}

In Figure~\ref{fig:correc_confid_p} we show an example of toy dataset and the relationship between confidence, correctness, as well as its multiplication. We can see that those metrics are correlated. Note, however, points may have confidence near 0.5 but correctness 0.0, that is, the model produce fairly high confidence for the correct class, but may be still insufficient to classify it correctly. On the other hand, the model may have also confidence near 0.5 and be able to correctly classify the example. Using only correctness may also be prone to errors, since some instance may be correctly classified in half of the iteratios, but still present low confidence in average, below 0.4. This motivated us to use the product instead of each metric independently. 

\begin{figure}[htbp]
\begin{center}
\begin{tabular}{cc}
\includegraphics[width=0.475\linewidth]{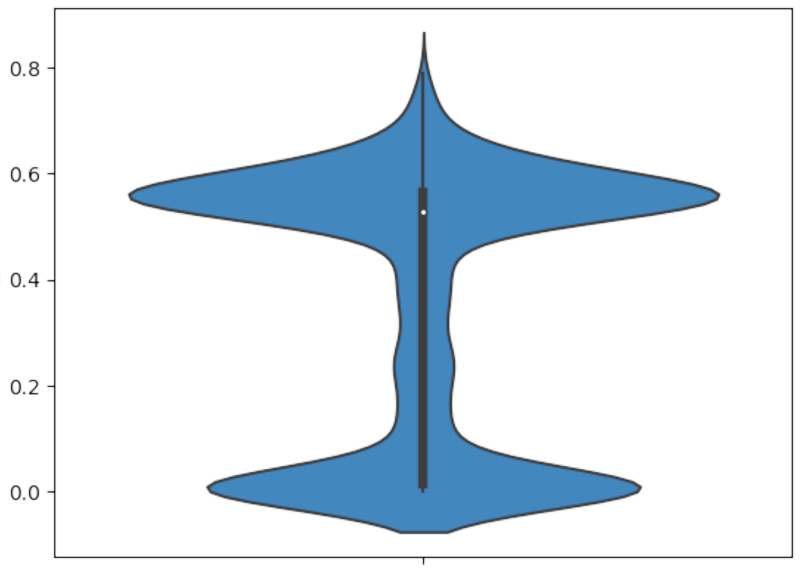}&
\includegraphics[width=0.475\linewidth]{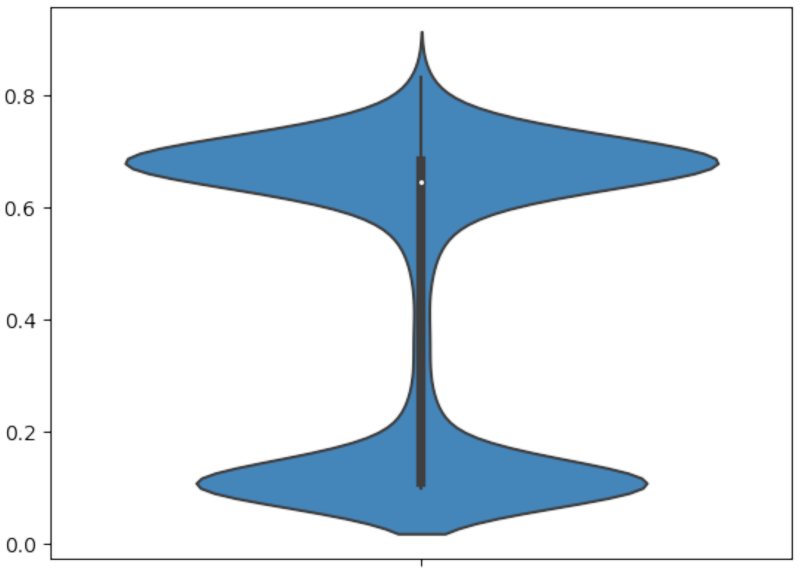}  \\
     (a) $m_j$ distribution & (b) $w_j$ distribution
\end{tabular}
\end{center}
\caption{Examples of violin plot with the distributions of (a) $m_j$ correctness and confidence product, and (b) $w_j$ learned weights across the "binary synthetic" dataset instances}
\label{fig:weights_distribution_threshold}
\end{figure}

We propose using the product between confidence and correctness as an interesting option to assess the difficulty of an example to be learned, as well as indicating a possible label noise. In this -- simple yet effective -- approach we use either a visual inspection on the distribution of $m_j$ across all values of the training set as illustrated in Figure~\ref{fig:weights_distribution_threshold}-(a), or use a validation set in order to find a threshold, in order to remove examples from the training set and retrain the model. Note from the violin plots that, while using the $m_j$ directly does show two evident modes that may be separated into noisy and not-noisy instances, there is a way of adjusting such weights to make this separation even more clear (see Figure~\ref{fig:weights_distribution_threshold}-(b)), as we detail in the next section.

\subsection{Learning weights for iterative noise detection}

Although a simple threshold may improve the results of a model by removing potential pathological instances, it may be difficult to find an optimal cutoff value. Therefore, our second approach learns the weights of each instance, reducing the weights of pathological ones iteratively during $E$ iterations. As detailed in Algorithm~\ref{algo:weights}, it starts by initializing all instances with equal weights, and for each iteration $l$ uses the correctness and confidence product to adjust such weights, using it in the training process of the GBDT algorithm. The hypothesis is that noisy values will have progressively lower weights when compared with clean instances.

\begin{algorithm}
\caption{Iterative weight learning for noise detection}
\label{algo:weights}
\begin{algorithmic}
\REQUIRE Training set: $X, Y$ with instances $j=1\ldots n$
\STATE $w_j \gets 1/n \text{, } j=1\ldots n$ \COMMENT{initialize instances weights}
\FOR {$l=0$ \TO $E$}
\STATE train GBDT $h(.)$ with $(X,Y)$ and weights $w_j$
\STATE $\mu_j, \sigma_j, c_j \gets$ DatasetCartography($X,Y,h$)
\STATE $w_j \gets w_j - (1- c_j\cdot \mu_j)$ \COMMENT{weights adjustment}
\STATE $w_j \gets \operatorname{clip}(w_j,0,1)$ \COMMENT{clip weights to range [0,1]}
\ENDFOR
\end{algorithmic}
\end{algorithm}

After learning the weights, it is possible to visually inspect the distribution, as shown in Figure~\ref{fig:weights_distribution_threshold}-
(b), or use other strategy to find the cutoff point for noise detection. Note that this approach creates a more well separated distribution of weights $w_j$ when compared with the direct use of $m_j$ product of correctness and confidence values, even allowing for automatic separation via a algorithms that find the valley of such distribution separating the modes. What we want is to separate the mode with the lowest weights, that indicate difficult or pathological instances.

Our method is heuristic by design. In practical data science, having the option to adjust the threshold has the advantage of allowing the data scientist or analyst to adjust the threshold following a sensitivity/sensibility compromise. This is paramount for decision making related to the business goals. Nevertheless note that our method still allows for automatically finding a threshold based on the distribution of the metrics $w_j$ or $m_j$, for example by searching for the valley of such distribution.

\section{Experiments}

\subsection{Code repository}

The scripts for computing the training dynamics metrics with GBTD, as well as the datasets used in our experiments (except for the closed dataset) are available at \url{https://github.com/mapontimeli/cartography_gbdt}. We also include visualizations of the datasets, and a complete report of different metrics.

\subsection{Datasets}
In this paper we investigate 5 datasets, 2 public, 2 synthetic and a closed one as a case study for the industry, as summarized in Table~\ref{tab:datasets}. Synthetic datasets and Real datasets with added noise are important to allow for controlled experiments. Also, the real datasets are standard datasets often used on papers about tabular data. Each dataset has a specific characteristic, which allows us to demonstrate how the algorithms behave under different scenarios.

\begin{table}[htbp]
\caption{Datasets characteristics}
\begin{small}
\begin{center}
\begin{tabular}{|c|c|c|c|c|c|}
\hline
Dataset & \#Instanc. & \#Feat. & \#Class. & Types & Real \\
\hline
Binary & 15100 & 2 & 2 & Num & No\\
Multiclass & 16500 & 2 & 4 & Num & No\\
Adult & 32561 & 14 & 2 & Cat/Num & Yes\\
Breast Cancer & 569 & 30 & 2& Num & Yes\\
Meli Items & 39567 & 1201 & 4 & Cat/Num & Yes\\
\hline
\end{tabular}
\label{tab:datasets}
\end{center}
\end{small}
\end{table}

Both synthetic datasets were build using the data generator toolkit from the Python library Scikit-Learn (version 1.1.2). The main goal was to create a 2-dimensional set of instances by composing clusters and complex shapes, like moons and circles, in an unbalanced proportion of labels.

The public datasets were chosen to investigate real world scenarios for tabular numerical data (Breast Cancer Wisconsin Diagnosis) and mix of numerical and categorical features (Adult).

Each set was split in train, validation and test subsets using a 80-10-10 proportion (except for the Breast Cancer dataset in which was used a proportion of 67-16.5-16.5 given it's reduced size) in a stratified fashion to keep the label distribution over the sets. 

\begin{figure}[htbp]
\centerline{\includegraphics[width=0.95\linewidth]{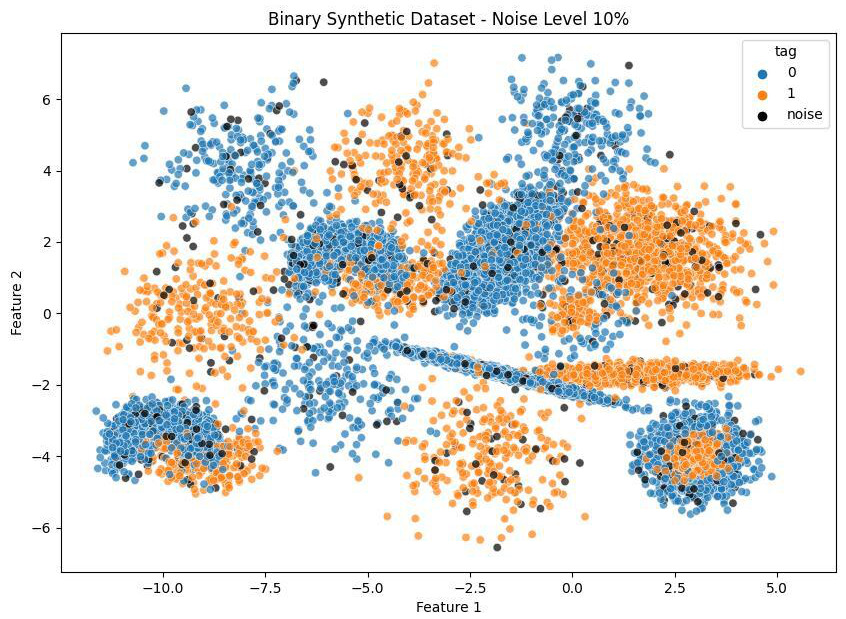}}
\centerline{\includegraphics[width=0.95\linewidth]{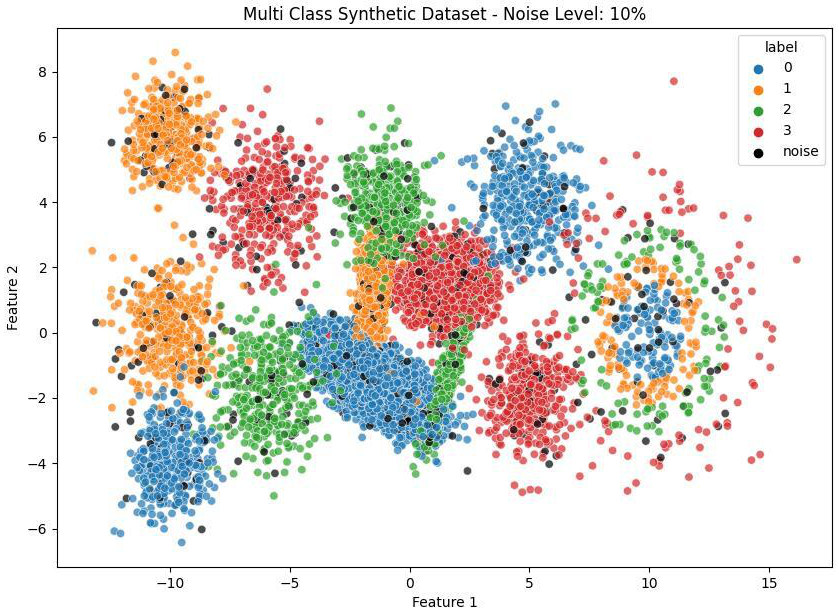}}
\caption{Synthetic datasets with 10\% noise, binary version (top) and multiclass version (bottom)}
\label{fig:dataset_synthetic}
\end{figure}

All raw datasets are assumed to be clean and the label noise was introduced artificially only to train and validation sets. 

Whenever we needed to use the datasets other than with LightGBM, e.g. to compute distances for neighbour search, or train an SVM classifier (for reasons we detail in the next sections), a one-hot encoding of categorical variables, as well as 0-1 scaling for numerical variables is performed.

\subsubsection{Noise Completely at Random (NCAR)}
We investigated three noise level proportions: 10\%, 20\% and 30\%. In Figure~\ref{fig:dataset_synthetic} we display the 10\% NCAR noisy synthetic datasets.

For binary classification problems the noise was simulated by flipping the labels randomly following the specified proportion. In the multi class ones, since those are unbalanced, we exchange labels but setting a limit in order to keep the label counting order of as in the original dataset.

\subsubsection{Noise Not at Random (NNAR)}
In this case, we use a neighbourhood constraint in order to select a pair of instances and exchange its labels. It is carried out by searching, within the $k$ nearest neighbors of each instance, those with different label, and randomly select one of those to exchange their labels with probability $p$. This process is repeated until we reach the desired amount of noise. As for the NCAR case, we generated versions of the datasets with 10, 20 and 30\% of label noise.

\subsection{Baselines and evaluation}

Our baselines are methods implemented in the libraries daal4py\footnote{\url{https://pypi.org/project/daal4py/}}, 
cleanlab\footnote{\url{https://github.com/cleanlab/cleanlab}} and doubtlab\footnote{\url{https://github.com/koaning/doubtlab}}. In daal4py we have the BrownBoost method, in cleanlab the Confident Learning approach~\cite{northcutt2021confident}, while doubtlab uses an ensemble of methods and heuristics. In particular we used an ensemble of the following reasons to ``doubt'' the quality of the label of a given instance:
\begin{itemize}
    \item Confident Learning;
    \item Low Probability: assign doubt when a models' confidence values are below a threshold of 0.55;
    \item Short Confidence, assign doubt when the correct label gains too little confidence, with threshold 0.05;
    \item Long Confidence, assign doubt when any wrong label gains too much confidence, with threshold 0.95;
    \item Disagreement, assign doubt when two models disagree on a prediction. In our case we used SVM and LightGBM. 
\end{itemize}

Our methods are based on training dynamics metrics, in particular:
\begin{itemize}
    \item threshold on $m_j$: is carried out by computing $m_j = c_j \cdot \mu_j$, i.e. correctness times confidence, and plot its distribution (as illustrated in Figure~\ref{fig:weights_distribution_threshold}), visually finding a threshold for each dataset;
    \item threshold on $w_j$: similar to the previous one, but the distribution is with respect to the learned weights $w_j$ after $E=10$ iterations.
\end{itemize}

\noindent \textbf{Evaluation} considers (i) compare the different methods and their ability to detect instances with incorrect/noisy label, and (ii) compare classification results of training with the noisy training set and training after removing the detected noisy labels. 

Let a positive instance be one with label noise. The following are used to evaluate the noise detection methods:
\begin{itemize}
    \item False Positive Rate: FPR = FP / (TN+FP);
    \item False Negative Rate: FNR = FN / (TP+FN).
\end{itemize}

We used the precision, recall, f1-score (macro average in the multiclass scenario) as well as the area under the precision/recall curve (PRAUC) to evaluate the binary classification.

\subsection{Experimental Setup}

Although our implementation is ready for both XGBoost and LightGBM (LGBM), only the latter is used as base classifier for brevity. Also, for each dataset and method, we used Optuna~\cite{akiba2019optuna} in order to tune the hyperparameters of the classifier, optimizing PRAUC for binary and f1-score macro average for multiclass ones for the validation set. 

This way, for each noise type (NCAR and NNAR) and noise level, 3 experiments were carried out. In the first we train (and optimize hyperparameters) with the noisy data. In the second we train (and optimize) after noise removal with training dynamics Hard-Threshold (thresh $m_j$), training dynamics Weigth-Threshold (thresh $w_j$), cleanlab, doubtlab. Thirdly we train BrownBoost, which is a robust model, using the noisy datasets directly. In all experiments with learned threshold weight $w_j$, the number of iterations is set to $E=10$, which was shown to be sufficient to converge. Some datasets require even less iterations for the weights to stabilize.

\section{Results and discussion}

\subsection{Synthetic Datasets}

\textbf{}
We first show the results of incorrect label (or noisy label) detection, detailed in Table~\ref{tab:res_noise_binary} for Binary dataset, and Table~\ref{tab:res_noise_multi} for the Multiclass dataset. 

It is possible to note the method based on a threshold on the learned weights $w_j$ present superior results on all synthetic results, however for multiclass with 10\% and 20\% noise level for which the threshold on $m_j$ is comparable with the learned weights. In both binary and multiclass casees, doubtlab heuristics performed worse, while cleanlab was competitive with noise type NCAR, but often performing badly for NNAR.

\begin{table}[]
\caption{Noise Detection Results for Binary Synthetic Dataset under NCAR and NNAR noise types}
\label{tab:res_noise_binary}
\begin{center}
\begin{tabular}{llrrrr}
             &               &     \multicolumn{2}{c}{NCAR}   &    \multicolumn{2}{c}{NNAR}                   \\
noise & method                & FPR    & FNR             & FPR & FNR\\ \hline\hline 
10\%         & thresh $m_j$   & 04.1 & 23.4     & 03.5     & 17.5       \\
             & thresh $w_j$   & 06.7 & 22.8       & 07.5     & 03.9              \\
             & cleanlab       & 03.0 & 18.3     &  03.3     & \em 73.4            \\
             & doubtlab       & 08.7     & 44.5     & 08.0     & \em 39.2            \\\hline
20\%         & thresh $m_j$   & 05.8     & 04.6     &  04.0     & \em 60.4       \\
             & thresh $w_j$   & 05.9 & 05.0 & 10.4     & 13.4              \\
             & cleanlab       & 03.4 & 13.7     & 09.9     & \em 83.2            \\
             & doubtlab       & 11.2     & 17.5     & 06.4     & 25.9            \\\hline
30\%         & thresh $m_j$   & 26.6     & 10.6     & 05.9     & \em 68.1       \\
             & thresh $w_j$   & 22.0 & 17.5      & 10.2     & 22.1              \\
             & cleanlab       & 05.5 & 26.2     & 01.4     & \em 89.2            \\
             & doubtlab       & 31.4     & 02.3     & 05.8     & 22.8            \\\hline

\end{tabular}
\end{center}
\end{table}

\begin{table}[]
\caption{Noise Detection Results for Multiclass Synthetic Dataset under NCAR and NNAR noise types}
\label{tab:res_noise_multi}
\begin{center}
\begin{tabular}{llrrrr}
             &               &     \multicolumn{2}{c}{NCAR}   &    \multicolumn{2}{c}{NNAR}                   \\
noise & method                & FPR    & FNR             & FPR & FNR\\ \hline\hline 
10\%         & thresh $m_j$   & 06.5 & 01.6             & 06.5 & 15.4      \\
             & thresh $w_j$   & 06.7 & 01.8             & 06.0 & 18.2              \\
             & cleanlab       & 02.1 & 09.0            &  02.1 &  \em 74.3          \\
             & doubtlab       & 07.3 & 22.8     & 06.5     & \em 59.7            \\\hline
20\%         & thresh $m_j$   & 06.5 & 02.1              & 01.9  & 65.5       \\
             & thresh $w_j$   & 06.7 &  01.9              & 01.6  & 68.9              \\
             & cleanlab       & 03.4 & 06.4             & 01.3  & \em 82.2  \\
             & doubtlab       & 10.2 & 02.0     & 02.3 & 23.8            \\\hline
30\%         & thresh $m_j$   & 14.1  & 16.1        & 05.9     & \em 58.1       \\
             & thresh $w_j$   & 09.8  & 13.4            & 02.9     & \em 65.3               \\
             & cleanlab       & 17.5 & 15.2         & 01.8 & \em 85.2   \\
             & doubtlab       & \em 73.8     & 00.2     & 02.7     &  19.7            \\\hline
       
\end{tabular}
\end{center}
\end{table}

In terms of classification results for the synthetic datasets, we compared the use of the noisy NNAR datasets with the cleaned dataset for training. In Table~\ref{tab:result_class_synthetic} we show the results for Binary, and in Table~\ref{tab:result_class_multisynthetic} for the Multiclass. For more clarity we omit the hard-threshold $m_j$ since $w_j$ threshold was slightly better, and omit and doubtlab results since it did not performed well on noise detection. 

A significant improvement is observed with the cleaned data in both binary and multiclass scenarios. In a few cases (Binary with 10\% and 20\% noise) the BrownBoost algorithm results was worse than training with LightGBM directly on the noisy data. Overall, the use of our method was better in terms of classification metrics.

\begin{table}[]
\caption{Classification results of Binary Synthetic datasets with the NNAR data and after cleaning with the weight-based method}
\label{tab:result_class_synthetic}
\begin{center}
\begin{tabular}{llrrrr}
Dataset &               &     precision  & recall & f1-score & prauc\\ \hline\hline
Binary & LGBM   &         0.77    &  0.92      & 0.84 &  0.85\\
noisy  10\%        & cleaned $w_j$   &         0.94   &   0.81    &  0.87 & 0.90 \\ 
       & cleanlab   &         0.80   &   0.94    &  0.86 & 0.88 \\ 
       & BrownBoost &         0.80   &   0.66    &  0.73 & 0.78 \\ \hline
Binary &  LGBM  &         0.62    &  0.85      & 0.72 & 0.77 \\
noisy  20\%        & cleaned $w_j$   &         0.82   &   0.84    &  0.83 & 0.87\\ 
       & cleanlab   &         0.87   &   0.81    &  0.84 & 0.87 \\ 
       & BrownBoost &         0.84   &   0.56    &  0.67 & 0.76 \\ \hline
Binary & LGBM   &         0.47    &  0.74      & 0.59 & 0.64 \\
noisy  30\%        & cleaned $w_j$   &         0.58   &   0.92    &  0.71 & 0.76 \\ 
       & cleanlab   &         0.55   &   0.92    &  0.67 & 0.73 \\ 
       & BrownBoost &         0.70   &   0.61    &  0.65 & 0.70  \\ \hline
\end{tabular}
\end{center}
\end{table}

\begin{table}[]
\caption{Classification results of Multiclass Synthetic datasets with the noisy data and after cleaning with the weight-based method}
\label{tab:result_class_multisynthetic}
\begin{center}
\begin{tabular}{llr}
Dataset &               &     f1-macro\\ \hline\hline
Multiclass & LGBM &       0.90 \\
noisy  10\%       & cleaned $w_j$   &       0.89 \\ 
       & cleanlab   &        0.91 \\ 
       & BrownBoost &       0.90 \\ \hline
Multiclass & LGBM &      0.82 \\
noisy  20\%       & cleaned $w_j$   &    0.85 \\ 
       & cleanlab   &     0.76 \\ 
       & BrownBoost &      0.84 \\ \hline
Multiclass & LGBM  &     0.76  \\  
noisy  30\%       & cleaned $w_j$   &  0.82 \\ 
       & cleanlab   &  0.64 \\ 
       & BrownBoost &  0.77 \\ \hline
\end{tabular}
\end{center}
\end{table}

\subsection{Real Datasets}

We first show the results of incorrect label (or noisy label) detection, detailed in Table~\ref{tab:res_noise_cancer} for Breast Cancer, and in Table~\ref{tab:res_noise_adult} for the Adult dataset. 

For the Breast cancer dataset all methods performed fairly well for NCAR, the threshold on $w_j$ was the most competitive, it was the best or second best method, achieving good results across different noise levels. We note that the proposed methods achieved, in general, a better compromise between FPR and FNR.

Considering the Adult dataset, which is the most challenging one, again our proposed methods were competitive across all noise levels, in particular using weights $w_j$ achieved the best FPR/FNR compromise, with cleanlab and doubtlab often excceeding one of those metrics by a large margin.

In general, the ensemble of heuristics implemented by doubtlab could not achieve good results overall, except of specific scenarios. For all noise levels, at least one the proposed methods were comparable or better than cleanlab, in particular for NNAR which is a more difficult noise type to handle.

\begin{table}[]
\caption{Noise Detection Results for Breast Cancer Dataset under NCAR and NNAR noise types}
\label{tab:res_noise_cancer}
\begin{center}

\begin{tabular}{llrrrr}
             &               &     \multicolumn{2}{c}{NCAR}   &    \multicolumn{2}{c}{NNAR}                   \\
noise & method                & FPR    & FNR            & FPR & FNR\\ \hline\hline 
10\%         & thresh $m_j$   & 05.2 & 07.8             & 01.1 & 50.0      \\
             & thresh $w_j$   & 04.7 &05.2         & 04.3 & 31.5              \\
             & cleanlab       & 03.1 & 10.6         &  02.0 &   50.0          \\
             & doubtlab       & 10.2 & 23.6     & 07.5     & 36.8            \\\hline
20\%         & thresh $m_j$   & 02.5 & 14.4         & 01.3  & 53.1       \\
             & thresh $w_j$   & 03.7 & 11.3             & 01.6  & 51.3              \\
             & cleanlab       & 07.8 & 09.2         & 00.9  & 48.1  \\
             & doubtlab       & 08.1 & 22.3     & 08.9      & 30.8            \\\hline
30\%         & thresh $m_j$   & 02.4  & 39.1    & 01.5     & 70.4       \\
             & thresh $w_j$   & 06.4  & 13.1            & 01.8  & 61.7               \\
             & cleanlab       & 02.9 & 11.4         & 03.8      & 61.8   \\
             & doubtlab       & 15.8     & 14.9     & 05.6     &  31.3           \\\hline
       
\end{tabular}
\end{center}
\end{table}

\begin{table}[]
\caption{Noise Detection Results for Adult Dataset under NCAR and NNAR noise types}
\label{tab:res_noise_adult}
\begin{center}
\begin{tabular}{llrrrr}
             &               &     \multicolumn{2}{c}{NCAR}   &    \multicolumn{2}{c}{NNAR}                   \\
noise & method                & FPR    & FNR            & FPR & FNR\\ \hline\hline 
10\%         & thresh $m_j$   & 05.2 & 07.8     & 16.5 & 40.0       \\
             & thresh $w_j$   & 04.7 & 05.2     & 14.0 & 40.1              \\
             & cleanlab       & 03.1 & 10.6     & 08.7 & 54.5            \\
             & doubtlab       & 16.2 & 40.6     & 33.5 & 13.0            \\\hline
20\%         & thresh $m_j$   & 02.5 & 14.4     & 09.1 & 32.6        \\
             & thresh $w_j$   & 03.7 & 11.3     & 09.0  & 30.0              \\
             & cleanlab       & 07.8 & 09.2     & 05.3  & 58.7  \\
             & doubtlab       & 08.1 & 22.3     & 22.2  & 40.6            \\\hline
30\%         & thresh $m_j$   & 02.4  & 39.1    & 18.6  & 34.9     \\
             & thresh $w_j$   & 06.4  & 13.1     & 18.9  & 35.6               \\
             & cleanlab       & 02.9 & 11.4      & 01.8  & 81.7   \\
             & doubtlab       & 15.8  & 14.9     & 17.8  & 56.8           \\\hline
       
\end{tabular}
\end{center}
\end{table}


We also compare classification results for the real datasets. In Table~\ref{tab:result_class_real} the weight learning method and confident learning (cleanlab) method are used to clean the data since those achieved the best results. We again compare those against BrownBoost. 

Significant improvement is observed with the cleaned data, in particular for Breast Cancer, while BrownBoost showed no significant difference with respect to LightGBM. However, in the Adult dataset, as we increase the noise levels, the problem becomes more difficult, and in this case the BrownBoost method shows advantages, in particular for the 20 and 30\% noise levels. For 10\% noise, the removal techniques increased the recall, but reduced the precision.

\begin{table}[]
\caption{Classification results of Real datasets with the NNAR data and after cleaning with the weight-based and competing methods}
\label{tab:result_class_real}
\begin{center}
\begin{tabular}{llrrrr}
Dataset &               &     precision  & recall & f1-score & prauc \\ \hline\hline
Cancer & LGBM    &         0.94    &  0.89      & 0.91 & 0.93 \\
noisy 10\%        & cleaned $w_j$   &         0.97   &   0.86    &  0.91 & 0.94 \\ 
       & cleanlab   &         1.00   &   0.82    &  0.91 & 0.95\\ 
       & BrownBoost &            0.94    &  0.89      & 0.91 & 0.93  \\        \hline
Cancer & LGBM    &         0.91    &  0.91      & 0.91  & 0.93\\
noisy 20\%        & cleaned $w_j$   &         1.00   &   0.78    &  0.88 & 0.93 \\ 
       & cleanlab   &         0.86   &   0.91    &  0.89 & 0.91 \\ 
       & BrownBoost &         0.94    &  0.91      & 0.93 & 0.94 \\        \hline
Cancer & LGBM     &         0.80    &  0.80      & 0.80 & 0.84 \\
noisy 30\%       & cleaned $w_j$   &         0.84   &   0.89    &  0.87 & 0.88 \\ 
       & cleanlab   &         0.78   &   0.83    &  0.81 & 0.84 \\ 
       & BrownBoost &         0.78   &   0.80    &  0.79 & 0.83 \\ 
       \hline       \hline
Adult & LGBM  &       0.77    &  0.60   &   0.68 & 0.73 \\
noisy 10\%       & cleaned $w_j$   &    0.58   &   0.87    &  0.71  & 0.76 \\ 
       & cleanlab   &         0.58   &   0.79    &  0.70  & 0.74 \\ 
       & BrownBoost &         0.76   &   0.60    &  0.67  & 0.73\\   \hline
Adult & LGBM  &     0.74    &  0.61   &   0.66 & 0.72 \\
noisy 20\%       & cleaned $w_j$   &     0.85   &   0.46    &  0.64 & 0.72 \\ 
       & cleanlab   &         0.74   &   0.61    &  0.66  & 0.72\\ 
       & BrownBoost &         0.73   &   0.61    &  0.67  & 0.72 \\  \hline
Adult & LGBM  &    0.66    &  0.63   &   0.64 & 0.69  \\  
noisy 30\%        & cleaned $w_j$   &    0.0   &   0.0    &  0.0  & 0.62\\ 
       & cleanlab   &         0.86   &   0.15    &  0.26  & 0.61 \\ 
       & BrownBoost &         0.66   &   0.63    &  0.64 & 0.69\\ \hline
\end{tabular}
\end{center}
\end{table}

\subsection{Industry Scenario: forbidden items dataset}

In Mercado Libre (Meli), one of the initiatives of the fraud prevention team is to deploy models to detect forbidden items before such items are made available in the marketplace. This protects users from unsafe or illegal products. However, as in many real world scenarios, such labels come from different sources such as denounces, manual revision, regular expression detection and hard rules. When retrieved from a large time window from tens of thousands of items or more, such labels are usually noisy.

\begin{figure}[htbp]
\centerline{\includegraphics[width=0.95\linewidth]{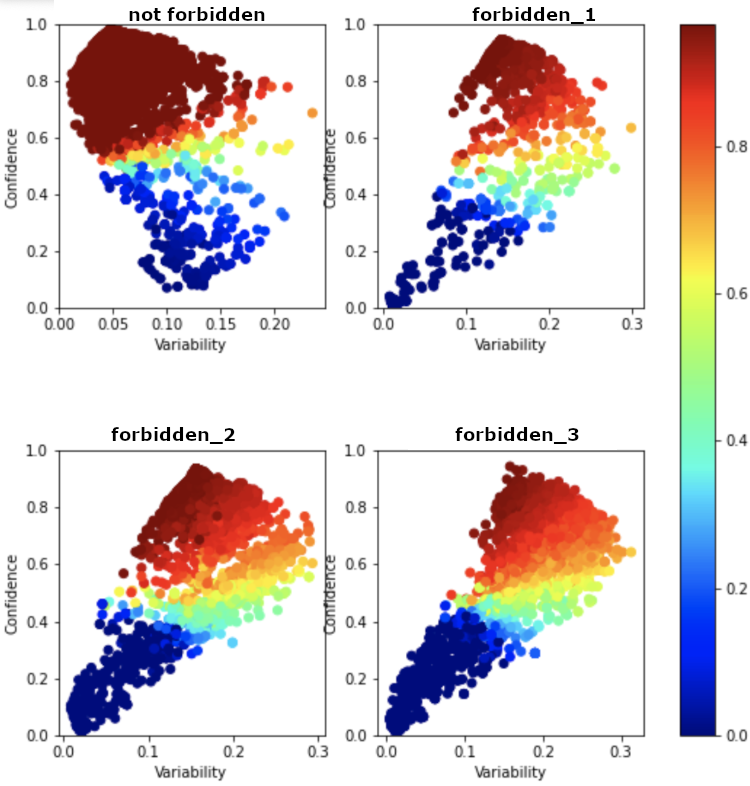}}
\caption{Cartography metrics for a forbidden item dataset with 4 classes. The colorbar represents correctness in the range [0,1].}
\label{fig:cartog_forbidden}
\end{figure}

Three categories of forbidden items are defined in this problem. As shown in Figure~\ref{fig:cartog_forbidden}, many instances of forbidden items have low correctness and confidence. We used the proposed method to clean the training set in order to avoid issues with conflicting labels. For example the item ``In-Ear Hearing Aid Power tone F-138'' and a similar (but not identical) version of it appeared in the dataset with both "not forbidden" and "forbidden" labels. Since it is a hearing device not approved by local regulators, it is forbidden. In contrast, some items related to ``Android Car Codec Player'' also showed different labels, but are not forbidden.

We used a combination of tabular features (from the item or its seller), as well as feature extracted from text, totalling 1201 features.

We used the proposed method, as well as the confident learning and doubtlab heuristics, to clean the training data removing the noisy examples. Additionally, we submitted a subset of examples for relabeling: with our method we selected a subsample of examples with weight zero (0.0) using the weights $w_j$, while all the noisy examples detected by the clean/doubtlab were relabeled. The results on a sample of productive data (200 items manually reviewed) are shown in Table~\ref{tab:res_forbidden}, along with the number of instances removed or submitted to relabeling. As confirmed by the simulated experiments, the method has strong potential to be used to map candidates for noisy items, but also to select instances for relabeling, significantly improving dataset quality.

After showing good results, such method is currently available to assess training sets, allowing data scientists to iterate versions of such dataset by either discarding instances, or submitting it to label validation. This fosters data quality and improves the model development cycle. 

\begin{table}[]
\caption{Classification results of Forbidden Item Meli Dataset comparing before and after cleaning and testing in productive data. We also detail the number of detected pathological instances to be cleaned or relabeled.}
\label{tab:res_forbidden}
\begin{center}
\begin{tabular}{lrrrr}
            &  \# path.inst & precision  & recall & f1-score \\ \hline\hline
original/noisy data & -- &   0.76    &  0.91      & 0.79 \\ \hline
cleaned $w_j$   &  437 &       0.82   &   0.88    &  0.84 \\
cleaned cleanlab   & 201 &        0.78   &   0.87    &  0.81  \\ 
cleaned doubtlab   & 159 &        0.71   &   0.80    &  0.75  \\ \hline
relabeled $w_j$      & 200 &       0.89   &   0.90    &  0.89 \\
relabeled cleanlab   & 201 &        0.86   &   0.87    &  0.85 \\ 
relabeled doubtlab   & 159 &        0.82   &   0.88    &  0.86 \\ \hline
\end{tabular}
\end{center}
\end{table}


\section{Conclusion}

Using training dynamics metrics is useful to detect pathological instances in noisy datasets. Our experiments showed superior performance when using the confidence and correctness metrics in order to iteratively separate such instances from clean ones using weights. In particular for cases with a high noise ratio, our experiments show significant improvement in f1-score and prauc when removing such instances from the training set. Going beyond the simulations, we also present a case study with a productive dataset, demonstrating the practical applicability of the method. We believe one of the reasons our method was better was to use the same algorithm (a GBTD) to assess the dataset and to train the model, removing instances that are hindering model learning process. 

Our main contribution is to present a methodology to assess datasets, in particular using ensembles of boosted decision trees, which are highly relevant methods in the industry since they allow to work with both categorical and numerical data. We show a practical scenario of application resulting in a deployed solution.

We believe more advances towards understanding and learning with such data under noise are still needed. Future work may investigate strategies for automatically relabeling the noisy candidate instances. Also, study other contexts of noisy labels such as class-dependent mislabeling, comparing this with GBDTs for particular scenarios can be interesting, as well as in other real world datasets.

\bibliographystyle{ACM-Reference-Format}

\bibliography{sample-base}

\end{document}